# Using stigmergy to incorporate the time into artificial neural networks

Federico A. Galatolo, Mario G.C.A. Cimino, and Gigliola Vaglini

Department of Information Engineering, University of Pisa, 56122 Pisa, Italy
{federico.galatolo, mario.cimino, gigliola.vaglini}@ing.unipi.it

**Abstract.** A current research trend in neurocomputing involves the design of novel artificial neural networks incorporating the concept of time into their operating model. In this paper, a novel architecture that employs stigmergy is proposed. Computational stigmergy is used to dynamically increase (or decrease) the strength of a connection, or the activation level, of an artificial neuron when stimulated (or released). This study lays down a basic framework for the derivation of a stigmergic NN with a related training algorithm. To show its potential, some pilot experiments have been reported. The XOR problem is solved by using only one single stigmergic neuron with one input and one output. A static NN, a stigmergic NN, a recurrent NN and a long short-term memory NN have been trained to solve the MNIST digits recognition benchmark.

## 1      Introduction and background

The use of Artificial Neural Networks (ANN) in engineering is rapidly increasing in pattern recognition, data classification and prediction tasks [1,2]. An important difference in recent generations of ANN is the fact that they try to incorporate the concept of time into their operating model. For example, in *spiking* NNs a temporal spike is used for mapping highly nonlinear dynamic models. Networks of spiking neurons are, with regard to the number of neurons, computationally more powerful than earlier ANN models [3]. Nevertheless, training such networks is difficult due to the non-differentiable nature of spike events [4]. Another relevant class of ANN which exhibits temporal dynamics is that of *recurrent* (cyclic) NNs (RNNs). Unlike feedforward (acyclic) NNs, recurrent NNs can use their internal state (memory) to process sequences of inputs, creating and processing memories of arbitrary sequences of input patterns [3]. A special class of recurrent NN is based on the *Long Short-Term Memory* (LSTM) unit, which is made of a cell, an input gate, an output gate and a forget gate. The cell remembers values over arbitrary time intervals and the three gates regulate the flow of information into and out of the cell [3]. However, the training process of recurrent NNs strongly depends on the set of constraints and regularizations used in the optimization process, resulting in a not entirely unbiased task with respect to the "how" [5].

This paper focuses on a novel architecture that employs *stigmergy* to incorporate the concept of time in ANN. Stigmergy is defined as an emergent mechanism for self-coordinating actions within complex systems, in which the trace left by a unit's action on some medium stimulates the performance of a subsequent unit's action [6]. It is a fundamental mechanism in swarm intelligence and multi-agent systems, but it also models individual interactions [6]. In neuroscience, Hebb studied this phenomenon in the biological brain, as a basis for modeling synaptic plasticity, i.e., the ability of synapses to strengthen or weaken over time, in response to increases or decreases in their coordinated activity [7]. According to Hebb's theory, synaptic plasticity is one of the important neurochemical foundations of learning and memory. Specifically, the Hebb's law states that, when an axon of cell A is near enough to excite cell B and repeatedly or persistently takes part in firing it, some growth process or metabolic change takes place in one or both cells such that A's efficiency, as one of the cells firing B, is increased. This is often paraphrased as "neurons that fire together wire together" [7]. Similarly, in the phenomenon of selective forgetting that characterizes memory in the brain, neural connections that are no longer reinforced will gradually lose their strength relative to recently reinforced ones. Accordingly, computational stigmergy can be used to increase (or decrease) the strength of a connection, or the activation level, of an artificial neuron when stimulated (or unused).

To our knowledge, this is the first study that proposes and lays down a basic framework for the derivation of stigmergic NNs. In the literature, computational intelligence research using stigmergy is focused on swarm and multi-agent systems coordination, and on computational optimization [6]. Although its high potential, demonstrated by the use of stigmergy in biological systems at diverse scales, the use of stigmergy for pattern recognition and data classification is currently poorly investigated. As an example, in [8] a stigmergic architecture has been proposed to perform adaptive context-aware aggregation. In [9] a multilayer architectures of stigmergic receptive fields for pattern recognition have been experimented for human behavioral analysis. The optimization process of both systems is carried out using the Differential Evolution.

In this paper, the dynamics of stigmergy is applied to weights, bias and activation threshold of a classical neural perceptron, to derive the stigmergic perceptron (SP). To train a NN made of SPs, each stigmergic layer can be formally transformed into an equivalent static MLP, by spatial multiplication of nodes, layers and progressive removal of internal temporal steps. Once the network is totally unfolded, the resulting net is much larger, and contains a large number of weights. However, this static NN can be efficiently transformed in a computational graph. As a consequence, a given stigmergic NN can be transformed into a computational graph and trained using backpropagation.

To appreciate the impressive computational power achieved by stigmergic neurons, in this paper two experiments are shown. In the first experiment, the XOR problem is solved by using only one single stigmergic neuron with one input and one output. In the second experiment a static NN, a stigmergic NN, a recurrent NN and a LSTM NN have been trained to solve the MNIST digits recognition benchmark.

The remainder of the paper is organized as follows. Section 2 discusses the architectural design of stigmergic NNs with respect to traditional NNs. Experiments are covered in Section 3. Finally, Section 4 summarizes conclusions and future work.

## 2  Architectural design

This section defines formally the design of a stigmergic NN. Without loss of generality it is introduced a pilot application, with suitable input and output coding. Fig. 1a shows an example of static binary image, made of $4 \times 5 = 20$ pixels, in the spatial input coding. The necessary number of inputs for a static NN is $N = 20$. Fig. 1b shows the same image in the temporal input coding. Considering $T = 5$ instants of time for providing the image row by row, the necessary number of inputs of a temporal NN is $N = 4$. In essence, the image is provided in terms of chunks over 5 subsequent instants of time. Once provided the last chunk, the temporal NN provides the corresponding output class. Therefore, a NN that incorporates the concept of time is able to process the image as a temporal series. In the training phase, the NN learns to adjust its internal states to follow the chunks dynamics, according to the input coding.

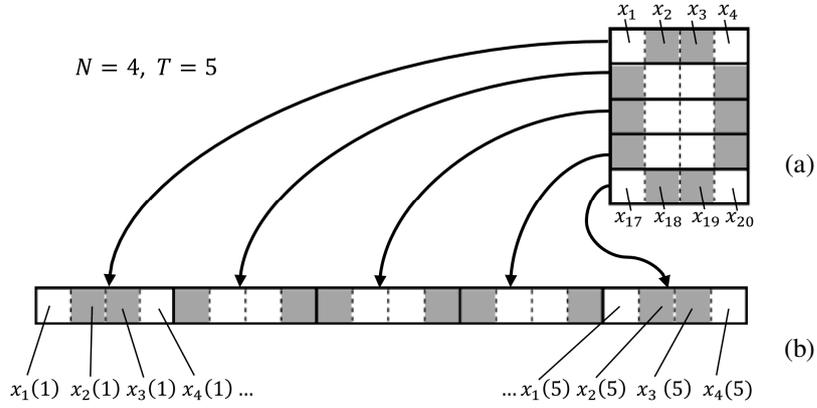

**Fig. 1.** (a) Spatial input coding. (b) Temporal input coding.

More formally, the conversion from spatial to temporal input coding can be represented as a spatial-to-temporal function $S2T(\cdot)$ providing a sequence of $T$ temporized chunks:

$$\langle x(1)|\ldots|x(T)\rangle = T2S(x) \qquad (1)$$

where $x = [x_1, \ldots, x_{N \cdot T}]$ and $x(t) = [x_1(t), \ldots, x_N(t)]$.

Fig. 2a and Fig. 2b show a conventional perceptron and a Stigmergic Perceptron (SP), respectively. The SP contains a smaller number of input connections, fed via temporal input coding; the weights $w_i(t)$ and the activation threshold $h(t)$ are dynamic. Note that the output value is provided only at time $T$, unless it is connected with a further SP.

More formally, let us assume a simple Heaviside step function in the activation gate without loss of generality. The conventional perceptron is modelled as follows:

$$u = w_0 + \sum_{1}^{N \cdot T} w_i \cdot x_i \qquad (2)$$

$$y = H(u) = \begin{cases} 0 & if\ u < h \\ 1 & if\ u \geq h \end{cases} \qquad (3)$$

whereas the SP processing is modelled as follows:

$$u(t) = w_0 + \sum_1^N w_i(t) \cdot x_i(t) \tag{4}$$

$$y(t) = H(u(t)) = \begin{cases} 0, & u(t) < h(t) \\ 1, & u(t) \geq h(t) \end{cases} \tag{5}$$

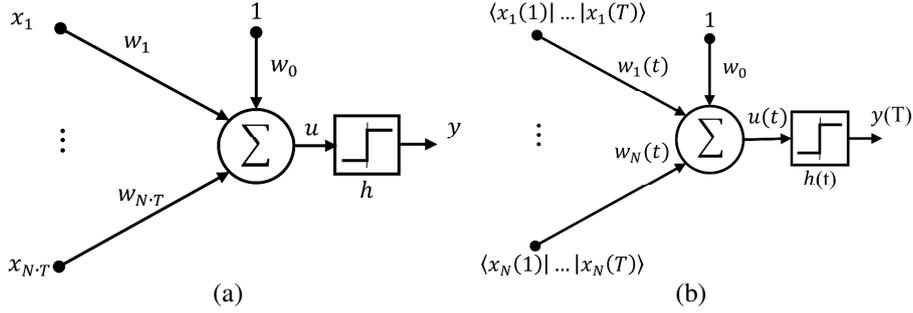

**Fig. 2.** (a) Conventional Perceptron (CP); (b) Stigmergic perceptron (SP).

The next step is to define the stigmergy dynamics ontologically and then formally in the operating model. Fig. 3a shows the stigmergy ontology [10], made by three concepts (or classes): *Time*, *Mark* and *Stimulus*. A *Stimulus* reinforces a *Mark* that, in turn, advances another *Stimulus*. The *Mark* can be reinforced up to a saturation level. Since the *Time* weakens the *Mark*, in absence of other *Stimulus* the *Mark* finishes. Fig. 3b shows the dynamics of a stigmergic relationship. On the bottom, a generic variable of class *Stimulus*: a binary variable generically called $s(t)$. On the top, a generic variable of class *Mark*: a real variable generically called $m(t)$, controlled by the stimulus variable. Specifically, the mark starts from $m(0)$ and, while the stimulus value is 0, undergoes a weakening by $\partial m$ per step, up to the minimum level $\underline{m}$. While the stimulus value is 1, the mark is reinforced by $\Delta m$ per step, up to the maximum level $\overline{m}$ of saturation.

More formally, the stigmergic relationship is defined as follows:

$$m(t) = \begin{cases} \max\{\underline{m}, m(t-1) - \partial m\}, & s(t, t-1, \ldots) = 0 \\ \min\{\overline{m}, m(t-1) - \partial m + \Delta m\}, & s(t, t-1, \ldots) = 1 \end{cases} \tag{6}$$

The mark dynamics can depend on current and previous values of the stimulus. According to Formula (6) the stigmergic relationships of the SP are:

$$w_i(t) = \begin{cases} \max\{\underline{w_i}, w_i(t-1) - \partial w_i\}, & x_i(t-1) = 0 \\ \min\{\overline{w_i}, w_i(t-1) - \partial w_i + \Delta w_i\}, & x_i(t-1) = 1 \end{cases} \tag{7}$$

$$h(t) = \begin{cases} \max\{\underline{h}, h(t-1) - \partial h\}, & y(t-1) = 0 \\ \min\{\overline{h}, h(t-1) - \partial h + \Delta h\}, & y(t-1) = 1 \end{cases} \tag{8}$$

In essence, the weight dynamics depends on the previous value of the input on the connection. The activation threshold depends on the previous value of the perceptron's output. The overall dynamics are parameterized via the initial mark value, the delta mark weakening, and the delta mark reinforcement. Thus, the training procedure will tune $m(0)$, $\partial m$, and $\Delta m$, i.e., $\{w_i(0)\}$, $\{\partial w_i\}$, $\{\Delta w_i\}$, $h(0)$, $\partial h$, and $\Delta h$.

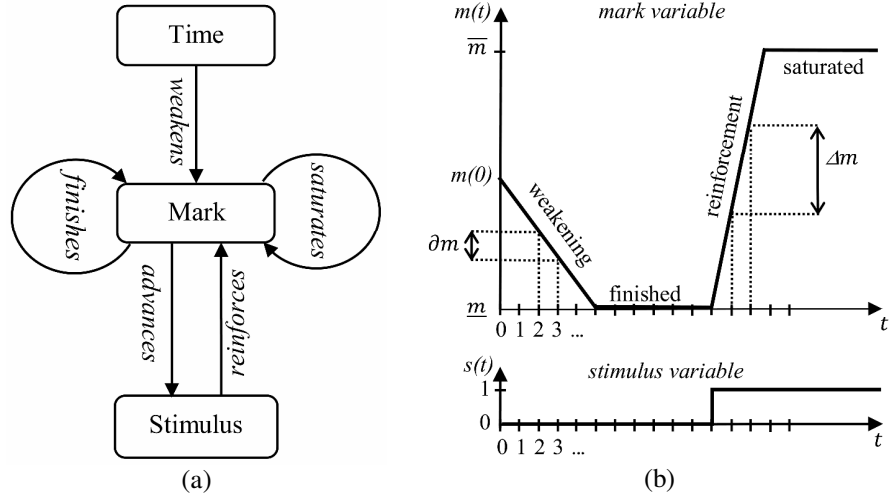

**Fig. 3.** (a) Ontology of stigmergy; (b) Dynamics of a stigmergic relationship.

Fig. 4a shows a fully-connected layer of SPs. It can be demonstrated that this stigmergic NN can be represented as the standard static NN shown in Fig. 4b, called unfolded equivalent NN. Specifically, in the unfolded NN each multilayer perceptron $MLP_i$ is the static equivalent of the $SP_i$, and receives all input values from 1 to $T$. The unfolding of a stigmergic NN is based on the progressive addition of nodes, layers for progressive removal of internal temporal steps, up to a completely static NN. Once the network is totally unfolded, the resulting net is much larger, and contains a large number of weights and layers.

The next step is to make the unfolded network differentiable. For this purpose, the step activation function can be approximated by a sigmoid function: $y = 1/(1 + e^{-(u-h)})$, where the midpoint $h$ corresponds to the soft threshold. Moreover, the mark variable can be modelled as a linear function with respect to the stimulus variable. Finally, the saturation/finishing constraints can be approximated by a sigmoidal clamping function applied to the mark variable.

At this step, the NN is then a static differentiable mathematical function, $\boldsymbol{y} = f(\boldsymbol{x})$, which can be efficiently represented in a computation graph (CG), the descriptive language of deep learning models [10].

When training a NN, the error is a function of the parameters. A CG is a functional description of the required computation, where to compute error derivatives with respect to all the parameters, for use in gradient descent, is very efficient. A CG can be instantiated for forward or backward computation. Specifically, reverse-mode differentiation, i.e. is the backpropagation, is carried out very efficiently.

As a consequence of the equivalence between a stigmergic NN and an unfolded NN, a given stigmergic NN can be transformed into a computational graph and efficiently trained using backpropagation.

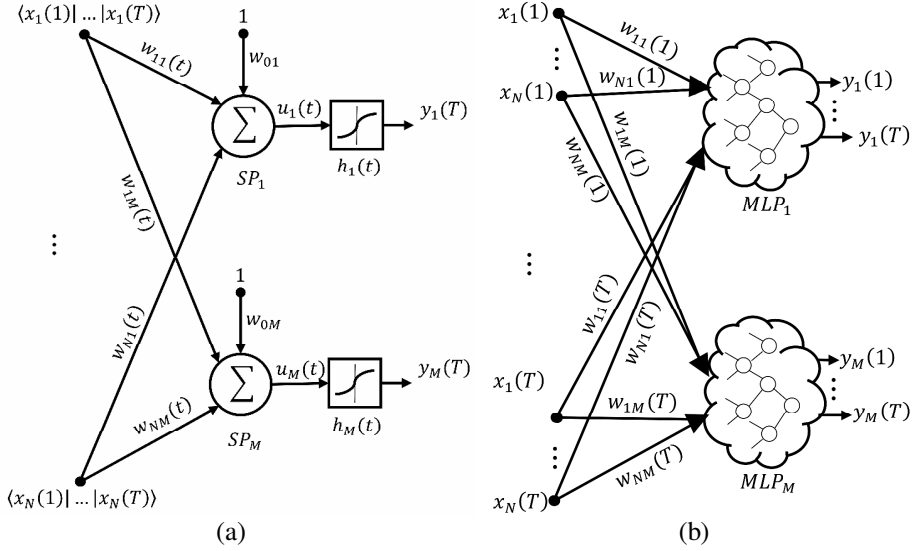

**Fig. 4.** (a) A fully-connected layer of SPs; (b) The unfolded equivalent NN.

## 3  Experimental studies

In order to appreciate the computational power of stigmergic neurons, in the first experiment the XOR problem is solved by using only one single stigmergic neuron, with one input and one output.

Assuming $w_0 = 0$ in Formula (4), and according to Formula (5):

$$y(0) = \begin{cases} 0, & w(0) \cdot x(0) < h(0) \\ 1, & w(0) \cdot x(0) \geq h(0) \end{cases} \qquad (9)$$

Assuming $\underline{w} = -\infty$, and $\overline{w} = \infty$, Formula (7) and Formula (8) become:

$$w(t) = w(t-1) - \partial w + x(t-1)\Delta w \qquad (10)$$
$$h(t) = h(t-1) - \partial h + y(t-1)\Delta h \qquad (11)$$

Thus, from Formula (5):

$$y(1) = \begin{cases} 1, (w(0) - \partial w + x(0)\Delta w)x(1) \geq h(0) - \partial h \text{ and } y(0) = 0 \\ 1, (w(0) - \partial w + x(0)\Delta w)x(1) \geq h(0) - \partial h + \Delta h \text{ and } y(0) = 1 \\ 0, \text{ else} \end{cases} \quad (12)$$

Fig. 5a shows a representation of y(0) and y(1), according to Formula (9) and Formula (12), in the (x(1), x(t+1)) space. Here, Formula (9) is represented by the dashed vertical line, whereas Formula (12) is represented by the two hyperbolas. In particular, four points are highlighted, and specified in Fig. 5b, where it is apparent that the XOR problem is solved: $\overline{y(1)} = x(0) \oplus x(1)$.

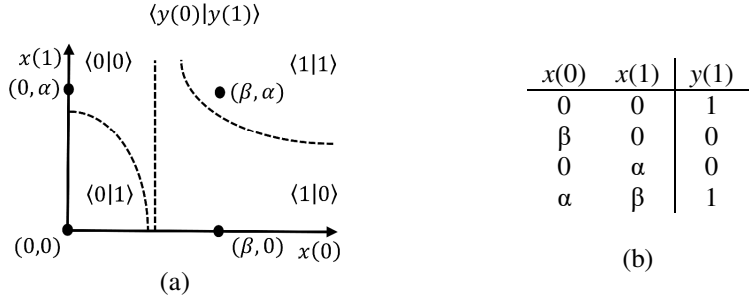

**Fig. 5.** The XOR problem with a single stigmergic neuron in the (x(1), x(t+1)) space.

In the second experiment, a static NN, a stigmergic NN, a recurrent NN and a LSTM NN have been trained to solve the MNIST digits recognition benchmark [10]. The purpose is twofold: to measure the computational power added by stigmergy to static perceptrons, and to compare the performances of existing temporal NN. First, the static NN and the stigmergic NN have been dimensioned to achieve the best classification performance. Subsequently, the other temporal NNs have been dimensioned to have a similar number of parameters with respect to the stigmergic NN.

Fig. 6 shows the architecture of the stigmergic NN. Here, the hourglass icon highlights the stigmergic relationships in the layer. Precisely, the 1st layer is a space-to-time coder (*S2T*); the 2nd layer is a set of fully connected perceptrons with stigmergic activation thresholds (*S$_h$LP*); the 3rd layer is a set of fully connected perceptrons with stigmergic weights and stigmergic activation thresholds *S$_{wh}$LP*; the 4th layer is a time-to-space decoder (*T2S*), which is the inverse transformation with respect to S2T; the 5th layer is a set of fully connected static perceptrons (*MLP*). The architecture of the static NN is made by two hidden layers of static perceptrons, and the output layer of 10 perceptrons.

The software has been developed with the PyTorch framework [13] and made publicly available on GitHub [14].

Precisely, an input image of the MNIST is made by 28×28 = 784 pixels, and the output is made by 10 classes corresponding to decimal digits. Overall, the data set is made of 70,000 images. At each run, the training set is generated by random extraction of 60,000 images; the remaining 10,000 images makes the testing set.

The static NN is made by 784 inputs connected to 300 perceptrons (235,200 weights and 300 biases) that, in turn, are connected to other 300 perceptrons (90,000 weights and 300 biases), that, in turn, are connected to 10 perceptrons (3000 weights and 10 biases), for a total number of parameters equals to 328,810 parameters.

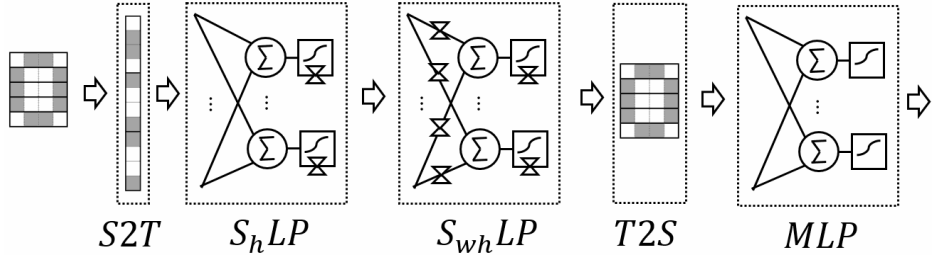

**Fig. 6.** A deep stigmergic NN for the MNIST digits recognition benchmark.

The stigmergic NN is made by 28 inputs connected to 10 perceptrons with stigmergic activation (280 weights, 10 biases, 10 initial activation thresholds, 10 delta thresholds weakening, and 10 delta thresholds reinforcement) that, in turn, are connected to 10 stigmergic perceptrons with weights and stigmergic activations (100 initial weights, 100 delta weight weakening, 100 delta weight reinforcement, 10 initial activation thresholds, 10 delta thresholds weakening, 10 delta thresholds reinforcement, and 10 biases), that become 280 output after the time-to-space decoder, which, in turn, are connected to 10 static perceptrons (2800 weights and 10 biases), for a total number of parameters equals to 3,470 parameters.

The Recurrent NN is fed by 28 inputs, organized identically to the stigmergic NN inputs. Such inputs are fully connected to two parallel feed forward layers (2·28·28 weights and 2·28 biases); in one of these layers, each output neuron has a backward connection to the inputs of both layers (28·56 weights); in the other layer, the outputs are connected to a further feed forward layer with 10 outputs (28·10 weights and 10 biases). The total number of parameters is 3,482.

The LSTM NN is fed by 28 inputs, organized identically to the stigmergic NN inputs. For each LSTM layer, the number of parameters is calculated according to the well-known formula $4 \cdot o \cdot (i+o+1)$, where $o$ and $i$ is the number of outputs and inputs, respectively.
The topology is made by a 28×10 LSTM layer, a 10×10 LSTM layer, a 10×10 LSTM layer, and a final 10×10 Feed Forward layer. Thus, the overall number of parameters is $4[10(28+10+1)) + 2 \cdot 10(10+10+1)] + (10 \cdot 10+10+10) = 3,360$.

Table 1 shows the overall performance and complexity. The performance evaluations are based on the 99% confidence interval of the classification rate (i.e., the ratio of correctly classified inputs to the total number of inputs), calculated over 10 runs. The complexity values correspond to the total number of parameters. The Adaptive Moment Estimation (Adam) method [12] has been used to compute adaptive learning rates for each parameter of the gradient descent optimization algorithms, carried out with batch method.

It is worth nothing that the MNIST benchmark is a spatial dataset, used to favor static NN and to show the concept of space-time mapping that can be exploited with temporal NNs. As a such, chunks sequences in each image are not inherently related in time. Nevertheless, the static NN employs a very large number of parameters, about two order of magnitude larger with respect to the temporal NNs.

On the other hand, Static NN, LSTM NN and Stigmergic NN have similar classification performances (differing only by 2% at the most). Comparing the classification rates of temporal NNs, which have been dimensioned to have a similar complexity, recurrent NN is largely outperformed by static NN and LSTM NN.

In consideration of the relative scientific maturity of the other comparative networks, the experimental results with the novel stigmergic NN looks very promising, and encourage further investigation activities for future work.

**Table 1.** Performance and complexity of different ANN solving the MNIST digits recognition benchmark.

| NN type | Complexity | Classification rate |
|---|---|---|
| Static NN | 328,810 | .951 ± 0.0026 |
| LSTM NN | 3,360 | .943 ± 0.011 |
| Stigmergic NN | 3,470 | .927 ± 0.016 |
| Recurrent NN | 3,482 | .766 ± 0.033 |

## 4  Conclusions

In this paper, the dynamics of computational stigmergy is applied to weights, bias and activation threshold of a classical neural perceptron, to derive the stigmergic perceptron (SP). An efficient methodology is proposed for training a multilayered NN made of SP. The methodology is based on the equivalence with static computational graphs, which can be trained using backpropagation optimization algorithms.

The effectiveness of the approach is shown via pilot experiments. Stigmergic perceptrons can be appreciated for their impressive computational power with respect to conventional perceptron. Moreover, stigmergic layers can be easily employed in deep NN architectures, and can provide performances similar to other relatively mature temporal NN, such as Recurrent NN and LSTM NN, on equal complexity.

## Acknowledgements

This work was partially carried out in the framework of the SCIADRO project, co-funded by the Tuscany Region (Italy) under the Regional Implementation Programme for Underutilized Areas Fund (PAR FAS 2007-2013) and the Research Facilitation Fund (FAR) of the Ministry of Education, University and Research (MIUR).

This research was supported in part by the PRA 2018_81 project entitled "Wearable sensor systems: personalized analysis and data security in healthcare" funded by the University of Pisa